\documentclass{article}

\usepackage[final, nonatbib]{neurips_2024}

\usepackage[utf8]{inputenc} 
\usepackage[T1]{fontenc}    
\usepackage{url}            
\usepackage{booktabs}       
\usepackage{amsfonts}       
\usepackage{nicefrac}       
\usepackage{microtype}      
\usepackage{xcolor}         

\usepackage{nccmath}  
\usepackage{bbm}
\usepackage{amsmath}
\usepackage{amssymb}
\usepackage{mathtools}
\usepackage{amsthm}

\usepackage{colortbl} 
\usepackage{xcolor} 
\usepackage[pdftex]{graphicx}

\usepackage[flushleft]{threeparttable} 
\usepackage{multirow}

\usepackage{wrapfig}

\newcommand{\tf}[1]{\mathbf{#1}}

\definecolor{Note_color}{rgb}{0.0, 0.0, 0.0}

\definecolor{lightyellow}{RGB}{255, 255, 204}
\definecolor{lightred}{RGB}{255, 204, 204}

\usepackage{pifont}
\newcommand\mynuma[1]{\ifcase#1 \or \ding{172}\or \ding{173}\or
  \ding{174}\or \ding{175}\or \ding{176}\or \ding{177}%
  \or \ding{178}\or \ding{179}\or \ding{180}\or \ding{181}\else *\fi\relax}

\newcommand\mynumb[1]{\ifcase#1 \or \ding{182}\or \ding{183}\or
  \ding{184}\or \ding{185}\or \ding{186}\or \ding{187}%
  \or \ding{188}\or \ding{189}\or \ding{190}\or \ding{191}\else *\fi\relax}

\newcommand{\METHOD}{AmoebaLLM}

\title{
\METHOD{}: Constructing Any-Shape Large Language Models for Efficient and Instant Deployment
}

\author{
  Yonggan Fu, Zhongzhi Yu\thanks{Contributed equally.} , Junwei Li\footnotemark[1] , Jiayi Qian\footnotemark[1] , Yongan Zhang, Xiangchi Yuan, \\
  \textbf{Dachuan Shi, Roman Yakunin, Yingyan (Celine) Lin} \\
  Georgia Institute of Technology \\
  \texttt{\{yonggan.fu, celine.lin\}@gatech.edu}\\
}

\begin{document}

\maketitle

\begin{abstract}

Motivated by the transformative capabilities of large language models (LLMs) across various natural language tasks, there has been a growing demand to deploy these models effectively across diverse real-world applications and platforms. However, the challenge of efficiently deploying LLMs has become increasingly pronounced due to the varying application-specific performance requirements and the rapid evolution of computational platforms, which feature diverse resource constraints and deployment flows. These varying requirements necessitate LLMs that can adapt their structures (depth and width) for optimal efficiency across different platforms and application specifications.  
To address this critical gap, we propose \METHOD{}, a novel framework designed to enable the instant derivation of LLM subnets of arbitrary shapes, which achieve the accuracy-efficiency frontier and can be extracted immediately after a one-time fine-tuning. In this way, \METHOD{} significantly facilitates rapid deployment tailored to various platforms and applications. Specifically, \METHOD{} integrates three innovative components: (1) a knowledge-preserving subnet selection strategy that features a dynamic-programming approach for depth shrinking and an importance-driven method for width shrinking; (2) a shape-aware mixture of LoRAs to mitigate gradient conflicts among subnets during fine-tuning; and (3) an in-place distillation scheme with loss-magnitude balancing as the fine-tuning objective.
Extensive experiments validate that \METHOD{} not only sets new standards in LLM adaptability but also successfully delivers subnets that achieve state-of-the-art trade-offs between accuracy and efficiency. Our code is available at \url{https://github.com/GATECH-EIC/AmoebaLLM}.

\end{abstract}
\section{Introduction}
\label{sec:intro}

The remarkable abilities and transformative impacts of large language models (LLMs)~\cite{touvron2023llama, llama3, team2024gemma, achiam2023gpt} have been paralleled by a growing interest in deploying them across a wide range of real-world applications and diverse platforms. However, given the rapid evolution of computational platforms and varying application-specific requirements, the challenge of deploying LLMs efficiently on various platforms with differing specifications has become more pronounced. This is because diverse platforms often feature different resource constraints and deployment flows, necessitating LLMs with varying structures and shapes (i.e., depth and width) to achieve maximized execution efficiency, as affirmed by our profiling in Sec. \ref{sec:profiling}.
Moreover, even the same platform may have varying requirements for LLMs' execution efficiency depending on factors such as on-device battery status.
These varying requirements demand a flexible framework capable of adapting to both the intrinsic hardware constraints and the extrinsic demands of diverse application scenarios.

Existing efficient LLM solutions~\cite{frantar2023sparsegpt,sun2023simple,ma2023llm,an2024fluctuation,kim2024shortened,xia2023sheared}, which primarily use model compression to bridge the gap between the resource constraints of the target device and the prohibitive complexity of LLMs, fail to fully address the above needs. This is because these solutions either focus on a single dimension of compression, resulting in limited efficiency improvements, or require a costly fine-tuning process for each target platform with its unique specifications. This strategy is particularly unscalable and inefficient for deploying widely used public LLMs like LLaMA~\cite{touvron2023llama, llama3}, where each user must compress and fine-tune the LLMs for their specific platform and application needs.

In light of this, it is highly desirable to develop a suite of LLMs designed such that compressed subnets of arbitrary shapes, which can achieve the accuracy-efficiency frontier without the necessity of individual fine-tuning, can be instantly extracted, thus allowing for immediate adaptation to the diverse needs of various platforms and applications. To achieve this, previous one-for-all training techniques~\cite{yu2020bignas,wang2021attentivenas,wang2021alphanet,gong2022nasvit,cai2019once}, which strategically sample subnets for joint training to deliver models with switchable complexity, are promising candidates. However, directly applying them to pre-trained LLMs would lead to failure due to the following challenges: \underline{(1)} their adopted subnet sampling strategies, which are dedicated to models trained from scratch, are not applicable to extensively pre-trained LLMs as informative and critical components that store useful knowledge are highly likely to be skipped; \underline{(2)} jointly fine-tuning different subnets on commonly adopted small-scale tuning datasets can easily cause severe gradient conflict~\cite{yu2020gradient,liu2021conflict} for LLM weights pre-trained on a large corpus, thus leading to poor performance of all subnets.

To address these challenges, we propose a framework called \METHOD{}, which endows a given LLM with the ability to instantly derive compressed subnets of arbitrary shapes that can achieve the accuracy-efficiency frontier. We achieve this through the development of three key components of \METHOD{}'s one-for-all fine-tuning scheme: a subnet selection strategy, a trainable adapter design, and a fine-tuning objective. Specifically, we summarize our contributions as follows:

\begin{itemize}
    \item We develop a framework called \METHOD{}, which grants a given LLM the capability to deliver subnets of arbitrary shapes that achieve state-of-the-art (SOTA) accuracy-efficiency trade-offs after a one-time fine-tuning. In this way, \METHOD{} can greatly facilitate rapid deployment across varying platforms and applications. This is achieved by integrating the following three key components to enable one-for-all fine-tuning.
    
    \item For extracting high-quality subnets with diverse shapes, we propose a knowledge-preserving subnet selection strategy that features dynamic programming (DP)-based depth shrinking and importance-driven width shrinking. This addresses the aforementioned challenge \underline{(1)} by preserving the encoded factual knowledge and reasoning capabilities of pre-trained LLMs.

    \item For the trainable adapter during one-for-all fine-tuning, we propose a shape-aware mixture of LoRAs (SMoL), which selects and combines a sparse set of LoRAs~\cite{hu2021lora} using a gating function that takes the subnet shape as input. This technique addresses the aforementioned challenge \underline{(2)} by mitigating gradient conflicts among subnets. More importantly, once the target subnet shape is determined based on the target platform/application at deployment time, all the selected LoRAs can be merged into the LLM weights, thus eliminating overhead.

    \item For the fine-tuning objective, we enhance the in-place distillation strategy~\cite{yu2020bignas,wang2021alphanet,gong2022nasvit} by integrating a loss-magnitude balancing scheme. This scheme is based on the observation that the loss magnitudes of subnets with different shapes in pre-trained LLMs are unbalanced, leading to a bias toward specific subnets and thus poor overall performance. Our proposed technique effectively addresses this issue and improves the performance of all subnets.

    \item Extensive experiments validate that our \METHOD{} can deliver LLMs with instantly serviceable subnets of any shape, each performing better or on par with SOTA efficient LLM solutions. Additionally, when leveraging \METHOD{} as a pure LLM compression framework to compress a given LLM to the target parameter, it can achieve new SOTA compression effectiveness, thanks to its subnet selection strategy.

\end{itemize}

\begin{figure}[htbp]
\centering
\includegraphics[width=0.99\linewidth]{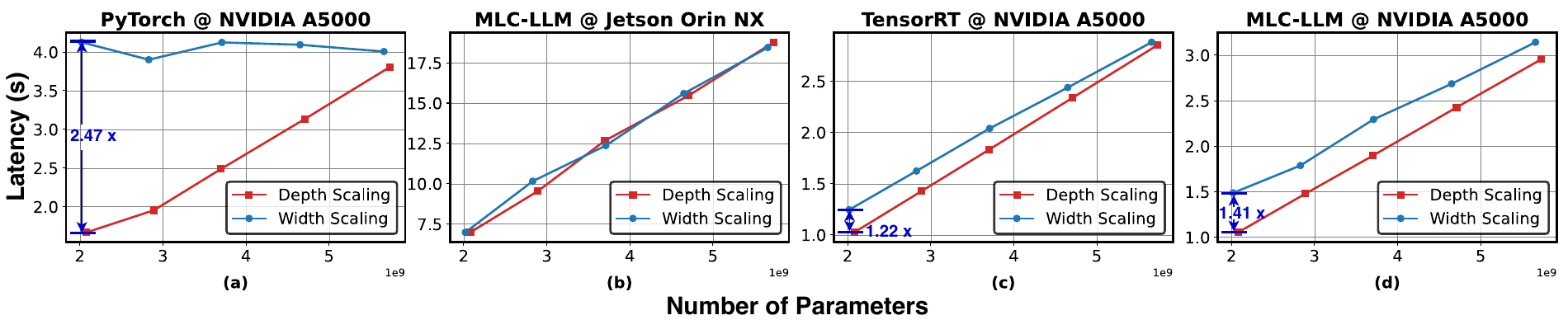}
\caption{The latency of LLaMA2 7B with scaled depth/width on various devices/deployment flows.}
\label{fig:profiling}
\vspace{-1em}
\end{figure}

\vspace{-0.5em}
\section{Motivation and Profiling}
\label{sec:profiling}
\vspace{-0.5em}

Before introducing our framework, we first conduct a profiling of generation latency across different devices, deployment flows, and use cases to examine the demand for LLMs with adaptable structures that meet the diverse needs of various platforms and applications.

\textbf{Profiling setup.} We study the efficiency of different LLM shapes on different devices by uniformly scaling either the depth or width of LLaMA2 7B~\cite{touvron2023llama} to the same model size. Here, depth is defined as the number of self-attention blocks, each including both a multi-head attention module and a feed-forward network, while width is defined as the hidden dimensions.
We profile these workloads using \underline{(1)} two devices, including an NVIDIA A5000 consumer-level GPU and an NVIDIA Jetson Orin NX edge GPU; and \underline{(2)} three deployment flows, including TensorRT-LLM~\cite{TensorRTLLM}, MLC-LLM~\cite{mlc-llm}, and vanilla PyTorch~\cite{paszke2019pytorch}.

\textbf{Observation and analysis.} As shown in Fig.~\ref{fig:profiling}, we can observe that \underline{(1)} first of all, the same workload on two GPUs with different resources exhibits large latency gaps, indicating the need for LLMs with adaptable structures to adapt to different devices when aiming to ensure a comparable latency to satisfy user needs; \underline{(2)} the preference regarding LLM shapes differs across deployment flows. Specifically, under the same model size, reducing model depth and width have a comparable impact on the measured latency using TensorRT-LLM on A5000 and MLC-LLM on Orin NX, while reducing depth using PyTorch or MLC-LLM on A5000 can achieve a notably lower latency than reducing width. This implies that for emerging platforms with limited compatible deployment flows, proper customization of LLM shapes for maximized efficiency is needed.

\section{The Proposed \METHOD{} Framework }
\label{sec:method}

\subsection{\METHOD{}: Methodology Overview}
\label{sec:method_overview}

\begin{figure}[t!]
\vspace{-1em}
\centering
\includegraphics[width=\linewidth]{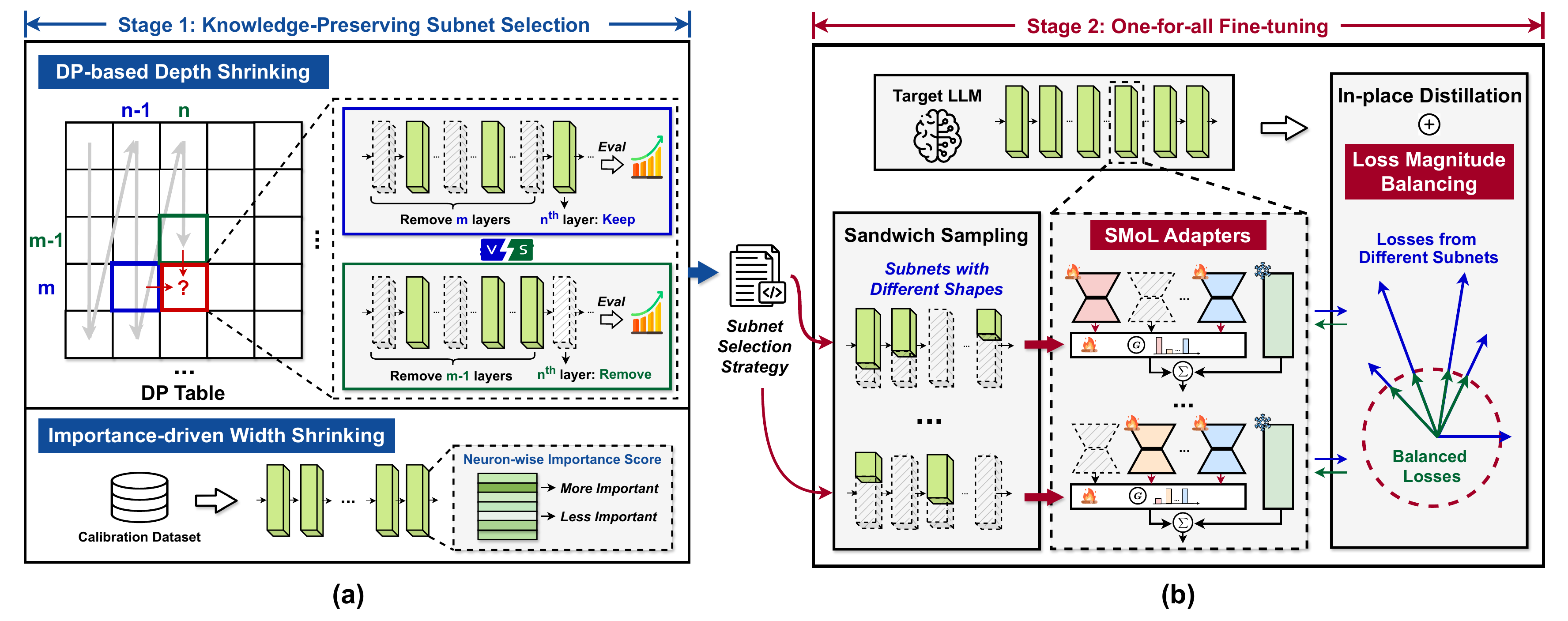}
\vspace{-1.5em}
\caption{An overview of our \METHOD{} framework: (a) Stage 1: Generate the subnet selection strategy; (b) Stage 2: One-for-all fine-tuning. Zoom in for a better view.}
\label{fig:overview}
\vspace{-1em}
\end{figure}

To address the challenges associated with traditional one-for-all network training, as mentioned in Sec.~\ref{sec:intro}, our \METHOD{} is equipped with three components: a knowledge-preserving subnet selection strategy, an SMoL adapter, and an in-place distillation fine-tuning objective with loss-magnitude balancing.
We illustrate our overall framework in Fig.~\ref{fig:overview}: given a target LLM, our \METHOD{} endows it with the capability of instantly deriving capable subnets via a two-stage process. 

In the first stage, \METHOD{} generates the subnet selection strategy. Specifically, given the target depth/width remaining ratios, this step decides which layers/neurons to maintain, respectively. To maximally preserve the knowledge and language modeling capabilities of pre-trained LLMs, we propose employing dynamic programming~\cite{bellman1954theory} to determine the retained layers under different remaining ratios and leverage neuron importance metrics~\cite{an2024fluctuation} to retain important neurons in a structured manner, as detailed in Sec.~\ref{sec:method_selection}. After this stage, the subnet selection strategy is determined and fixed.

In the second stage, we insert our proposed SMoL adapter into the target LLM for a one-time, one-for-all fine-tuning. Specifically, SMoL is composed of a set of LoRAs \cite{hu2021lora} and employs a gating function to sparsely activate different subsets of LoRAs for different subnets, thus mitigating their gradient conflicts. This process is elaborated upon in Sec.~\ref{sec:method_adapter}. In addition, our fine-tuning objective enhances the sandwich sampling and in-place distillation methods described in~\cite{yu2020bignas,wang2021alphanet,gong2022nasvit} by adding a loss-magnitude balancing scheme. This prevents bias towards specific subnets in pre-trained LLMs, as detailed in Sec.~\ref{sec:method_objective}. At deployment time, the selected LoRAs, which are determined by the extracted subnet shape favorable to the target platform, can be merged into the LLM weights.

\subsection{\METHOD{}: The Proposed Knowledge-Preserving Subnet Selection Strategy}
\label{sec:method_selection}

\textbf{Motivation.}
As detailed in Sec.~\ref{sec:related_work}, previous one-for-all training techniques \cite{yu2020bignas,wang2021attentivenas,wang2021alphanet,gong2022nasvit,cai2019once} often select the first layers of a model or the first channels of a layer. These techniques, intended for models trained from scratch, are unsuitable for pre-trained LLMs with rich knowledge encoded in their weights. Considering the difficulty of recovering lost knowledge through fine-tuning or our one-for-all fine-tuning on a relatively small corpus \cite{gekhman2024does}, it is crucial to identify informative and critical layers/neurons that store useful knowledge during the subnet selection process instead of relying solely on fine-tuning. To this end, we propose our knowledge-preserving subnet selection strategy to select the most informative layers/neurons under a given remaining ratio, as detailed below.

\textbf{DP-based depth shrinking.} 
Previous works have made diverse observations regarding the layer locations that store knowledge in different series of language models~\cite{zhang2024comprehensive, dai2021knowledge, geva2020transformer, jawahar2019does, meng2022locating, ju2024large}. As such, it is highly desirable to have a principled strategy to evaluate the joint contributions of different layer combinations in a target LLM to derive the optimal layer selection strategy for each remaining ratio. To achieve this goal, we propose a DP-based depth shrinking strategy.

\textit{Problem formulation.} 
Given a target LLM with \(N\) decoder layers, we define the layer selection strategy by a vector \(s \in \{0,1\}^{N}\). Here, \(s[n] = 1\) indicates that the \(n\)-th layer is retained; otherwise, the layer is removed. The objective is to determine the selection strategy \(s\) that achieves the optimal target metric, such as maximal accuracy or minimal perplexity (PPL), on a calibration dataset $C$, subject to the constraint that \(M\) layers are removed.

\textit{Key hypothesis.} Thanks to the residual structure \cite{he2016deep} of common LLMs \cite{touvron2023llama, llama3, team2024gemma} and the observations that LLMs' knowledge is compositional across layers \cite{zhang2024comprehensive, yao2023editing, wang2023easyedit}, we hypothesize that the layer selection problem described above can be divided into smaller and \textit{approximately} independent sub-problems. Consequently, we can employ dynamic programming~\cite{bellman1954theory} to effectively and efficiently solve the layer selection problem.

\textit{Our DP-based methodology.} We define a DP table \( \tf D[n][m] \) (where \( n \in [1, N] \) and \( m \in [1, M] \)), which stores the best target metric on the calibration dataset when exactly \( m \) layers are removed from the \textbf{\textit{first}} \( n \) layers of the target LLM. The corresponding layer selection strategy is denoted as \( \tf S[n][m] \in \{0,1\}^N \). Consequently, \( \tf S[N][M] \) represents the final strategy derived for removing \( M \) layers out of all \( N \) layers. Next, we elaborate on how to obtain \( \tf D[n][m] \) and \( \tf S[n][m] \), where we assume the target metric is such that larger values are better, without losing generality.

As illustrated in Fig.~\ref{fig:overview} (a), similar to general DP problems \cite{bellman1954theory}, \(\tf D[n][m]\) can be derived by a recurrence relationship. Specifically, to derive each \(\tf D[n][m]\), we compare the metrics achieved by the following two cases: \underline{(1)} removing \(m\) layers from the first \(n-1\) layers, and \underline{(2)} removing \(m-1\) layers from the first \(n-1\) layers and removing the \(n\)-th layer. The strategy yielding better metrics is adopted. More formally, this process can be formulated as follows:

\vspace{-0.5em}
\begin{equation}
    \tf D[n][m] = \max \left( \tf D[n-1][m], \tf P(n, m) \right) 
    \label{eq:recurrence}
\end{equation}
\noindent where \( \tf P(n, m) \) is the metric obtained by removing the \( n \)-th layer on top of the best-known strategy \( \tf S[n-1][m-1] \) for removing \( m-1 \) layers from the first \( n-1 \) layers. This is computed as follows:

\vspace{-0.5em}
\begin{equation}
    \tf P(n, m) = \textit{evaluate}(\textit{remove}(\tf S[n-1][m-1], n), C)
    \label{eq:update_strategy}
\end{equation}
where \(\textit{remove}(s, n)\) is a function that sets the \(n\)-th layer to 0 in a strategy \(s\).
Leveraging this recurrence relationship, after initializing the DP table with the base cases, i.e., \(\tf D[i][0] = \infty \) and \(\tf S[i][0] = \{1\}^N\) (\(\forall i \in [1, N]\)), since no layer is removed, the full DP table can be established with a complexity of \(\mathbb{O}(MN)\). In practice, we set \(M\) as the maximum number of layers allowed to be removed in our design space. Therefore, after obtaining the corresponding DP table \(\tf D\), the layer configuration of any subnet with \(m\) layers removed (\(\forall m \in [1, M]\)) can be directly obtained from \(\tf D[N][m]\). 
Note that constructing the DP table is a one-time effort for a given LLM.

\textit{Differences from and connections with previous methods.} The most relevant direction is layer pruning for LLMs. Pioneering works along this direction either utilize single-layer importance to determine which layers to prune \cite{kim2024shortened}, failing to measure the layers' joint contribution and thus aggressively losing factual knowledge as shown in Sec.~\ref{sec:exp}, or rely on pre-defined rules \cite{gromov2024unreasonable}, i.e., removing consecutive layers among the last ones, which is suboptimal and may not be generalizable to future LLMs. In contrast, our DP-based strategy can measure the joint impact of different layer combinations and is principled and fully automated without relying on human priors for specific LLMs. \textit{More importantly}, the two aforementioned works \cite{kim2024shortened,gromov2024unreasonable} are subsets of our DP-based strategy’s solution space. Thus, it can serve as a new SOTA layer pruning method, as demonstrated in Sec.~\ref{sec:exp_component_ablation}, in addition to serving as a component (i.e., the subnet selection strategy) in one-for-all fine-tuning. 
Previous works~\cite{lee2020flexible,xu2023efficient} have also applied dynamic programming for pruning at different granularities with varied formulations, and our work is the first to introduce this classical approach to LLMs.

\textbf{Importance-driven width shrinking.} Compared to layer selection, removing neurons (and all corresponding weight connections) for width shrinking is more fine-grained and involves a much larger design space. Therefore, instead of using dynamic programming, we directly employ existing structured pruning metrics for LLMs to obtain importance scores for each neuron and select the most important ones during one-for-all training, given a width remaining ratio.

To achieve this, we employ the metric in~\cite{an2024fluctuation} due to its outstanding performance compared to other alternatives. Specifically, we adopt the same width remaining ratio for linear layers within the same block (either the self-attention block or the feed-forward network), where the importance scores of input neurons of the last linear layer in each block are used to determine the width configuration of this block. In other words, when a subset of input neurons to the last linear layer is removed, all associated neurons and weights~\cite{ma2023llm} within this block will be removed in a structured manner.

\textit{Importance metric.} The importance score \( \mathbf{F}^{\ell}_{i} \) of the \(i\)-th input neuron in the last linear layer \( \mathbf{W}^{\ell} \) of the \(\ell\)-th block is computed as 
\( \mathbf{F}^{\ell}_{i} = \underset{k, t, j}{\mathbb{E}} ( \mathbf{X}^{\ell}_{k,t,i} - \overline{\mathbf{X}}^{\ell}_{i} )^2 \cdot \| \mathbf{W}^{\ell}_{j,i} \|_2^2 \), where \( j \) is the index of the output neurons, \( \mathbf{X}^{\ell}_{k,t,i} \) is the input features of the \( t \)-th token in the \( k \)-th batch, and \( \overline{\mathbf{X}}^{\ell}_{i} \) is the averaged input features across these two dimensions, both received by the \( i \)-th input neuron. To better maintain the LLMs' capability under a given remaining ratio: (1) the importance score is further normalized over all input neurons in each layer and globally sorted for non-uniform width shrinking; (2) a pre-computed bias term \( \mathbf{B}^{\ell} \) is added to the output neurons to compensate for the removed input neurons, i.e., 
\( \mathbf{B}^{\ell} = \mathbf{W}^{\ell} ( (1 - \mathbf{M}^{\ell}) \odot  \overline{\mathbf{X}}^{\ell}) \), where \( \mathbf{M}^{\ell} \) is the binary mask indicating whether the input neurons are retained.
We refer the readers to~\cite{an2024fluctuation} for more details.

\subsection{\METHOD{}: The Proposed Shape-aware Mixture of LoRAs}
\label{sec:method_adapter}

\textbf{Motivation.}
As demonstrated in Sec.~\ref{sec:exp_component_ablation}, joint weight fine-tuning of different subnets on small-scale datasets can lead to severe gradient conflicts~\cite{yu2020gradient, liu2021conflict}, resulting in poor performance across all subnets. One potential solution is to adopt parameter-efficient adapters like LoRA~\cite{hu2021lora}. However, accumulating gradients from all subnets onto the same LoRA still suffers from gradient conflicts, making fine-tuning unstable and causing some subnets to underperform. On the other hand, tuning a separate LoRA for each subnet configuration is infeasible due to the large design space. To address this, we propose an intermediate solution to balance performance and efficiency: the SMoL adapter that features a set of LoRAs, which are sparsely activated based on the subnet shape.

\textbf{SMoL adapter design.} Our SMoL adapter consists of a set of \(T\) LoRAs \(\{\mathbf{\Delta W}_{i} = \tf B_{i} \tf A_{i}\}_{i=1}^T\), which are sparsely activated for each subnet shape using a gating function \(\mathbf{G}\). Specifically, we employ a one-hot mask \(\mathbf{M}\) to indicate the shape, i.e., the layer/width configuration, of the subnet. This mask is fed into the gating function \(\mathbf{G}\) to calculate a score for each LoRA. Only the top \(k\) LoRAs are activated and weightedly averaged for each subnet shape during fine-tuning, thus mitigating the gradient conflicts among different subnets.

\textbf{Implementation.}
We extend the noisy top-K gating mechanism from \cite{shazeer2017outrageously} to implement our SMoL. Specifically, \(\mathbf{G}(\mathbf{M}) = \text{Softmax}(\text{KeepTopK}(\mathbf{H}(\mathbf{M}), k))\), where \(\text{KeepTopK}(\mathbf{v}, k)_i\) is \(v_i\) if \(v_i\) is in the top \(k\) of \(\mathbf{v}\) and \(-\infty\) otherwise, and \(\mathbf{H}(\tf x)_i = (\tf x \cdot \mathbf{W}_g)_i + \textit{std} \cdot \text{Softplus}((\tf x \cdot \mathbf{W}_{\text{noise}})_i)\). Here, \(\mathbf{W}_g\) and \(\mathbf{W}_{\text{noise}}\) are learnable, with the latter used to control the noise \(\textit{std}\) for load balancing \cite{shazeer2017outrageously}.
The final composited weight \(\mathbf{W} = \mathbf{W}_{base} + \sum_{i=1}^{T} \mathbf{G}(\mathbf{M})_i \mathbf{\Delta W_{i}}\), where \(\mathbf{W}_{base}\) is the pre-trained weight.

\textbf{Key difference from previous works.} Previous mixture-of-LoRA designs \cite{wu2023mole,chen2024llava,li2024mixlora} are input-adaptive at inference time, and thus the weights of different LoRAs cannot be merged into the original model. In contrast, our SMoL depends only on the subnet shape and is independent of model inputs. This implies that once the target subnet is extracted based on the target platform, all activated LoRAs can be merged into the model weights, thus enhancing parameter efficiency at deployment time.

\subsection{\METHOD{}: The Fine-tuning Objective}
\label{sec:method_objective}

\textbf{Motivation.}
Strategically sampling and jointly fine-tuning different subnets is necessary to ensure high-performance subnets within the same LLM, where the sandwich sampling and in-place distillation mechanisms~\cite{yu2020bignas,wang2021alphanet,gong2022nasvit} can serve as promising sampling and fine-tuning schemes, respectively. However, naively doing so can cause the larger subnets to gradually underperform during fine-tuning. We identify that this is due to notable differences in the loss magnitudes of different subnets, with smaller subnets, which diverge more from the well pre-trained full model, exhibiting much higher losses than larger subnets. To address this, we propose a simple but effective solution: equipping the in-place distillation with a loss-magnitude balancing mechanism, which we elaborate on as follows.

\textbf{In-place distillation with loss-magnitude balancing.} 
During each fine-tuning iteration, we employ the sandwich sampling~\cite{yu2020bignas,wang2021alphanet,gong2022nasvit} to sample \(K\) subnets \(\{\mathcal{T}_{i}\}_{i=1}^{K}\) with different layer/width remaining ratios, including the largest/smallest ones and \(K-2\) random ones from our design space. Detailed layer/width configurations of sampled subsets can be obtained from the strategies derived in Sec.~\ref{sec:method_selection}. We fine-tune our SMoL adapter as detailed in Sec.~\ref{sec:method_adapter} by accumulating the gradients from all sampled subnets using in-place distillation, where only the loss of the largest subnet $\mathcal{T}_{1}$ is calculated using ground truth, while those of other subnets \(\{\mathcal{T}_{i}\}_{i=2}^{K}\) use distillation from the largest one~\cite{yu2020bignas}. To balance the loss magnitude from different subnets, we normalize all subnets' loss magnitudes to that of the largest subnet, as visualized on the rightmost side of Fig.~\ref{fig:overview} (b). In this way, the final loss direction is jointly determined by all subnets' loss directions, falling on a unisphere without being severely impacted by their unbalanced magnitudes. We formulate the  fine-tuning objective as follows:

\vspace{-0.5em}
\begin{equation}
    \mathcal{L}_{\text{total}} = \mathcal{L}_{\text{CE}}(\mathcal{T}_{1}(x), y) +  \sum_{i=2}^{K} \frac{\|\mathcal{L}_{\text{CE}}(\mathcal{T}_{1}(x), y)\|}{\| \mathcal{L}_{\text{CE}}(\mathcal{T}_{i}(x), \mathcal{T}_{1}(x)) \|}  \mathcal{L}_{\text{CE}}(\mathcal{T}_{i}(x), \mathcal{T}_{1}(x))
    \label{eq:objective}
\end{equation}
where \(x\) and \(y\) are the input and ground truth, respectively, and \(\mathcal{L}_{\text{CE}}\) is a cross-entropy loss function.

\textbf{Final subnet search after fine-tuning}.
We adopt a simple hierarchical search strategy to select subnets from the fine-tuned LLM to satisfy the target efficiency constraint while maximizing achievable accuracy. Specifically, we first perform a coarse grid search across uniformly spaced depth and width settings based on a small calibration set (e.g., 40 samples from the MMLU dataset) to identify subnets that meet the efficiency constraint with maximized accuracy. Next, we conduct a more fine-grained grid search within depth and width ranges surrounding the optimal subnet identified in the coarse grid search stage. More advanced search strategies, such as evolutionary search~\cite{yu2020bignas,wang2021attentivenas,wang2021alphanet,gong2022nasvit}, are left for future work.

\section{Experimental Results}
\label{sec:exp}

\subsection{Experiment Setup}

\textbf{Baselines.} Our baselines include two SOTA structured width pruning methods: LLM-Pruner~\cite{ma2023llm} and FLAP~\cite{an2024fluctuation}, and one layer pruning method Shortened LLaMA~\cite{kim2024shortened}. All these baselines are open-sourced and we apply their official code to different LLMs. All baselines, including FLAP which were not fine-tuned in their original paper, are fine-tuned using the settings below for a fair comparison.

\textbf{Fine-tuning setting.}
Following~\cite{ma2023llm, kim2024shortened}, we adopt 50K samples from Alpaca~\cite{alpaca} for our one-for-all fine-tuning as well as for fine-tuning all baselines. For both our method and the baselines, we adopt a constant learning rate of 2e-4 with an AdamW optimizer and a LoRA rank of 64, and fine-tune for 10K iterations. It takes 40 GPU hours on an NVIDIA A5000 GPU for our one-for-all fine-tuning.

\textbf{Models.} We apply our method to LLaMA2 7B~\cite{touvron2023llama} and Vicuna 7B v1.5~\cite{vicuna}.

\textbf{Evaluation.}
Following our baselines~\cite{ma2023llm, kim2024shortened, an2024fluctuation, sun2023simple}, we leverage lm-evaluation-harness~\cite{eval-harness} to measure the zero-shot accuracy on 7 commonsense reasoning datasets, including BoolQ~\cite{clark-etal-2019-boolq}, PIQA~\cite{Bisk2020piqa}, HellaSwag~\cite{zellers2019hellaswag}, WinoGrande~\cite{sakaguchi2019winogrande}, ARC-easy~\cite{clark2018think}, ARC-challenge~\cite{clark2018think}, and OpenbookQA~\cite{OpenBookQA2018}. We also benchmark on the factual knowledge dataset MMLU~\cite{hendrycks2020measuring}.

\textbf{\METHOD{}'s setting.}  
We set the depth choices to range from 20 to 32 and the width remaining ratios as \{1, 7/8, 3/4, 5/8, 1/2\} by default unless specifically stated. In each iteration, we sample 4 subnets for joint training and select 2 LoRAs out of a total of 5 LoRAs in SMoL for each subnet.

\subsection{Benchmark with SOTA LLM Compression Methods}
\label{sec:exp_benchmark}

We benchmark our \METHOD{} against SOTA LLM width/layer pruning methods on LLaMA2 7B/Vicuna 7B v1.5 in Tab.~\ref{tab:benchmark_sota}/Tab.~\ref{tab:vicuna}, respectively. Note that the subnets produced by \METHOD{} are instantly extracted from the same one-for-all fine-tuned LLM, where the (depth, width scale) settings for 80\%/65\%/50\% remaining ratios, determined by the final subnet search in Sec.~\ref{sec:method_objective}, are (30, 0.875)/(28, 0.75)/(22, 0.75), respectively.
We also provide the per-subnet fine-tuned counterparts of our delivered subnets, denoted as \METHOD{}$^{\dagger}$, to compare one-for-all and individual fine-tuning.

\begin{table}[!t]
\centering
\vspace{-1.5em}
\caption{Compare with baseline methods under varying remaining ratios on LLaMA2 7B. }
\label{tab:benchmark_sota}
\resizebox{\linewidth}{!}{
\begin{tabular}{c|c|c|ccccccccc}\toprule
Ratio &\textbf{Method} $\downarrow$ &\textbf{MMLU} $\uparrow$ &\textbf{Average} $\uparrow$ &\textbf{BoolQ} $\uparrow$ &\textbf{PIQA} $\uparrow$ &\textbf{HellaSwag} $\uparrow$ &\textbf{WinoGrande} $\uparrow$ &\textbf{ARC-e} $\uparrow$ & \textbf{ARC-c} $\uparrow$ &\textbf{OBQA} $\uparrow$ \\\cmidrule{1-11}
\multirow{5}{*}{80\%} &LLM-Pruner~\cite{ma2023llm} &29.63 &56.95 &58.53 &76.39 &65.80 &60.38 &64.60 &34.56 &38.40 \\
&FLAP~\cite{an2024fluctuation} &40.21 &60.98 &\textbf{73.64} &74.81 &68.27 &65.43 &66.20 &37.88 &40.60 \\
&Shortened LLaMA~\cite{kim2024shortened} &26.45 &58.72 &62.17 & 76.01 & 68.22 & 58.88 & 68.98 & 38.40 & 38.40 \\\cmidrule{2-11}
&\cellcolor{lightyellow} \METHOD{} (Ours) &\cellcolor{lightyellow} 40.70 &\cellcolor{lightyellow} 62.29 &\cellcolor{lightyellow} 72.70 &\cellcolor{lightyellow} \textbf{76.80} &\cellcolor{lightyellow} 70.60 &\cellcolor{lightyellow} \textbf{67.60} & \cellcolor{lightyellow} 68.30 &\cellcolor{lightyellow} 38.80 &\cellcolor{lightyellow} \textbf{41.20} \\
&\cellcolor{lightred} \METHOD{}$^{\dagger}$ (Ours) &\cellcolor{lightred} \textbf{42.40} &\cellcolor{lightred} \textbf{62.37} &\cellcolor{lightred} 72.50 &\cellcolor{lightred} 76.30 &\cellcolor{lightred} \textbf{70.80} &\cellcolor{lightred} 66.90 & \cellcolor{lightred} \textbf{70.30} &\cellcolor{lightred} \textbf{40.20} &\cellcolor{lightred} 39.60 \\ \midrule \midrule
\multirow{5}{*}{65\%} &LLM-Pruner~\cite{ma2023llm} &23.15 &54.09 &60.73 &\textbf{74.97} &58.57 &57.85 &55.68 &33.02 &37.80 \\
&FLAP~\cite{an2024fluctuation} &33.28 &56.12 &65.75 &70.08 &60.57 &61.33 &62.25 &33.87 &\textbf{39.00} \\
&Shortened LLaMA~\cite{kim2024shortened} &24.89 &52.57 &62.32 & 72.03 & 55.10 & 52.41 & 59.47 & 30.63 & 36.00 \\\cmidrule{2-11}
&\cellcolor{lightyellow} \METHOD{} (Ours) &\cellcolor{lightyellow} 36.00 &\cellcolor{lightyellow} 56.96 &\cellcolor{lightyellow} \textbf{72.10} &\cellcolor{lightyellow} 70.70 &\cellcolor{lightyellow} 59.70 &\cellcolor{lightyellow} \textbf{63.20} &\cellcolor{lightyellow} 62.20 &\cellcolor{lightyellow} 34.00 &\cellcolor{lightyellow} 36.80 \\
&\cellcolor{lightred} \METHOD{}$^{\dagger}$ (Ours) &\cellcolor{lightred} \textbf{36.20} &\cellcolor{lightred} \textbf{57.26} &\cellcolor{lightred} 70.50 &\cellcolor{lightred} 70.90 &\cellcolor{lightred} \textbf{61.50} &\cellcolor{lightred} 62.70 &\cellcolor{lightred} \textbf{63.50} &\cellcolor{lightred} \textbf{34.50} &\cellcolor{lightred} 37.20 \\ \midrule \midrule
\multirow{5}{*}{50\%} &LLM-Pruner~\cite{ma2023llm} &22.90 &47.52 &61.83 &\textbf{67.79} &43.31 &51.22 &46.13 &28.16 &34.20 \\
&FLAP~\cite{an2024fluctuation} &27.67 &51.12 &59.45 &67.30 &51.33 &56.75 &55.43 &\textbf{31.57} &\textbf{36.00} \\
&Shortened LLaMA~\cite{kim2024shortened} &24.76 &47.35 &62.23 & 66.00 & 43.60 & 51.54 & 50.63 & 26.45 & 31.00 \\\cmidrule{2-11}
&\cellcolor{lightyellow} \METHOD{} (Ours) &\cellcolor{lightyellow} 30.60 &\cellcolor{lightyellow} 52.19 &\cellcolor{lightyellow} \textbf{65.70} &\cellcolor{lightyellow} 66.10 &\cellcolor{lightyellow} 51.30 &\cellcolor{lightyellow} 60.10 &\cellcolor{lightyellow} 56.60 &\cellcolor{lightyellow} 31.50 &\cellcolor{lightyellow} 34.00 \\
&\cellcolor{lightred} \METHOD{}$^{\dagger}$ (Ours) &\cellcolor{lightred} \textbf{32.20} &\cellcolor{lightred} \textbf{52.63} &\cellcolor{lightred} 64.70 &\cellcolor{lightred} 66.70 &\cellcolor{lightred} \textbf{53.00} &\cellcolor{lightred} \textbf{60.30} &\cellcolor{lightred} \textbf{58.00} &\cellcolor{lightred} 30.10 &\cellcolor{lightred} 35.60 \\
\bottomrule
\end{tabular}
}
\end{table}

\begin{table}[t!]\centering
\caption{Compare with baseline methods under varying remaining ratios on Vicuna 7B v1.5.}
\label{tab:vicuna}
\resizebox{\linewidth}{!}{
\begin{tabular}{c|c|c|ccccccccc}\toprule
\textbf{Ratio} &\textbf{Method} $\downarrow$ &\textbf{MMLU} $\uparrow$ &\textbf{Average} $\uparrow$ &\textbf{BoolQ} $\uparrow$ &\textbf{PIQA} $\uparrow$ &\textbf{HellaSwag} $\uparrow$ &\textbf{WinoGrande} $\uparrow$ &\textbf{ARC-e} $\uparrow$ &\textbf{ARC-c} $\uparrow$ &\textbf{OBQA} $\uparrow$  \\\cmidrule{1-11}
\multirow{5}{*}{80\%} &LLM-Pruner &38.94 &57.80 &64.27 &75.35 &64.28 &61.88 &64.35 &35.07 &39.40 \\
&FLAP &43.50 &60.54 &72.45 &73.67 &66.59 &65.98 &68.14 &37.54 &39.40 \\
&Shortened LLaMA &35.27 & 60.93 &67.58 & \textbf{75.68} & 68.12 & 64.48 & \textbf{70.20} & 40.44 & 40.00 \\\cmidrule{2-11}
&\cellcolor{lightyellow} \METHOD{} (Ours) & \cellcolor{lightyellow} 47.40 &\cellcolor{lightyellow} 60.77 &\cellcolor{lightyellow} 71.30 &\cellcolor{lightyellow} 72.70 &\cellcolor{lightyellow} \textbf{68.80} &\cellcolor{lightyellow} \textbf{66.10} &\cellcolor{lightyellow} 66.50 &\cellcolor{lightyellow} 38.60 &\cellcolor{lightyellow} \textbf{41.40} \\
&\cellcolor{lightred} \METHOD{}$^{\dagger}$ (Ours) &\cellcolor{lightred} \textbf{48.30} & \cellcolor{lightred} \textbf{61.54} &\cellcolor{lightred} \textbf{73.10} &\cellcolor{lightred} 73.33 &\cellcolor{lightred} 68.30 &\cellcolor{lightred} 65.80 &\cellcolor{lightred} 69.30 &\cellcolor{lightred} \textbf{40.80} &\cellcolor{lightred} 40.20 \\ \midrule \midrule
\multirow{5}{*}{65\%} &LLM-Pruner &24.07 &54.24 &61.19 &\textbf{73.50} &57.67 &57.85 &58.67 &32.42 &\textbf{38.40} \\
&FLAP &38.08 &56.47 &68.96 &70.78 &58.97 &61.33 &\textbf{63.43} &34.21 &37.60 \\
&Shortened LLaMA &25.59 &52.50 &64.28 & 70.62 & 56.78 & 57.46 & 52.74 & 31.40 & 34.20 \\\cmidrule{2-11}
&\cellcolor{lightyellow} \METHOD{} (Ours) & \cellcolor{lightyellow} 40.30 &\cellcolor{lightyellow} 55.93 &\cellcolor{lightyellow} 68.20 &\cellcolor{lightyellow} 69.80 &\cellcolor{lightyellow} 58.60 &\cellcolor{lightyellow} 62.40 &\cellcolor{lightyellow} 62.20 &\cellcolor{lightyellow} 34.10 &\cellcolor{lightyellow} 36.20 \\
&\cellcolor{lightred} \METHOD{}$^{\dagger}$ (Ours) &\cellcolor{lightred} \textbf{44.60} &\cellcolor{lightred} \textbf{56.74} &\cellcolor{lightred} \textbf{71.50} &\cellcolor{lightred} 69.60 &\cellcolor{lightred} \textbf{60.30} &\cellcolor{lightred} \textbf{64.50} & \cellcolor{lightred} 61.30 &\cellcolor{lightred} \textbf{34.40} &\cellcolor{lightred} 35.60 \\ \midrule \midrule
\multirow{5}{*}{50\%} &LLM-Pruner &23.24 &47.98 &59.08 &\textbf{68.55} &44.24 &52.17 &49.03 &28.41 &34.40 \\
&FLAP  &29.92 &50.74 &56.15 &67.52 &\textbf{51.81} &57.69 &56.57 &\textbf{31.06} &34.40 \\
&Shortened LLaMA &25.03 &45.56 & 51.71 & 65.67 & 43.28 & 51.38 & 49.96 & 26.54 & 30.40 \\\cmidrule{2-11}
&\cellcolor{lightyellow} \METHOD{} (Ours) &\cellcolor{lightyellow} 34.00 &\cellcolor{lightyellow} 51.41 &\cellcolor{lightyellow} 64.20 &\cellcolor{lightyellow} 65.00 &\cellcolor{lightyellow} 50.90 &\cellcolor{lightyellow} 59.10 &\cellcolor{lightyellow} 56.20 &\cellcolor{lightyellow} 30.90 &\cellcolor{lightyellow} 33.60 \\
&\cellcolor{lightred} \METHOD{}$^{\dagger}$ (Ours) &\cellcolor{lightred} \textbf{35.90} &\cellcolor{lightred} \textbf{52.36} &\cellcolor{lightred} \textbf{64.70} &\cellcolor{lightred} 65.70 &\cellcolor{lightred} 51.80 &\cellcolor{lightred} \textbf{60.80} &\cellcolor{lightred} \textbf{57.10} &\cellcolor{lightred} 31.00 &\cellcolor{lightred} \textbf{35.40} \\
\bottomrule
\end{tabular}
}
\vspace{-1em}
\end{table}

\textbf{Benchmark under comparable model sizes.}
As shown in Tab.~\ref{tab:benchmark_sota} and Tab.~\ref{tab:vicuna}, we observe that \underline{(1)} the subnets instantly extracted by \METHOD{} can achieve higher MMLU accuracy compared to all baselines, suggesting that our method better preserves the factual knowledge acquired during pre-training, as further analyzed in Sec.~\ref{sec:exp_component_ablation} and Sec.~\ref{sec:exp_calibration_dataset}; \underline{(2)} \METHOD{}'s delivered subnets, extracted from the same model, can also achieve better or comparable average commonsense reasoning accuracy compared to the strongest baselines, each trained separately; \underline{(3)} \METHOD{}$^{\dagger}$ achieves the best performance across all metrics and tasks compared to the baselines, indicating that \METHOD{}$^{\dagger}$ can serve as a new SOTA LLM compression framework in addition to its one-for-all functionality, thus advancing the achievable accuracy-efficiency trade-off; \underline{(4)} compared to our per-subnet fine-tuned variant \METHOD{}$^{\dagger}$, the instantly delivered subnets achieve comparable performance. This demonstrates the effectiveness of our one-for-all fine-tuning scheme, as further ablated in Sec.~\ref{sec:exp_component_ablation}.

\begin{wrapfigure}{r}{0.5\linewidth}
    \centering
    \includegraphics[width=\linewidth]{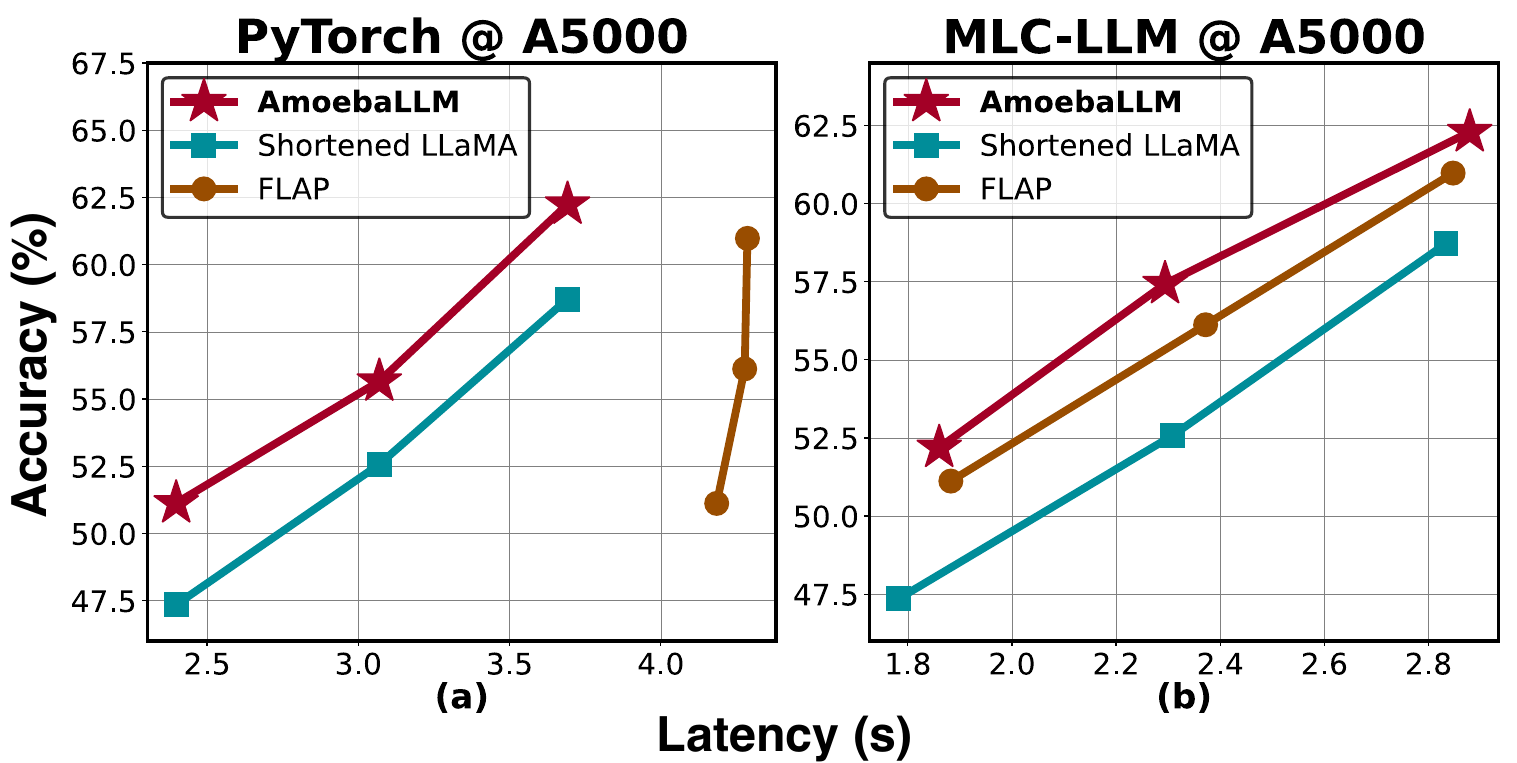}
    \vspace{-2em}
    \caption{Benchmark AmoebaLLM's achieved accuracy-latency trade-offs with SOTA LLM compression methods on an NVIDIA A5000 GPU.}
    \label{fig:measurement}
    \vspace{-0.5em}
\end{wrapfigure}

\textbf{Benchmark accuracy-latency trade-offs on real devices.}
We further benchmark the achieved trade-off between average commonsense reasoning accuracy and measured latency of LLaMA2 7B using MLC-LLM and PyTorch as the deployment flows on an NVIDIA A5000 GPU, following the settings in Sec.~\ref{sec:profiling}. For our method, we select the subnet shape that favors the hardware characteristics based on the profiling in Sec.~\ref{sec:profiling} from the one-for-all fine-tuned LLM. As shown in Fig.~\ref{fig:measurement}, we observe that \underline{(1)} our method consistently achieves the best trade-off on both deployment scenarios; and \underline{(2)} although FLAP achieves higher accuracy than Shortened LLaMA under comparable model sizes, its real-device speed is limited when using vanilla PyTorch, aligning with observations in~\cite{kim2024shortened}. In contrast, our method can instantly deliver subnets that favor the hardware/deployment flow characteristics, thus enjoying both high accuracy and real-device friendliness.

\vspace{-0.5em}
\subsection{Ablation Study: Effectiveness of Each Component}
\label{sec:exp_component_ablation}

We perform ablation studies to validate the effectiveness of each component of \METHOD{}.

\textbf{The DP-based depth shrinking strategy.}
We benchmark our DP-based strategy against two existing LLM layer pruning methods, Shortened LLaMA~\cite{kim2024shortened} and Unreasonable~\cite{gromov2024unreasonable}, on LLaMA2 7B. Specifically, we employ the three methods to select important layers using Wikitext2/MMLU with PPL/accuracy as calibration metrics, respectively. We directly report the achieved PPL/MMLU accuracy after calibration under various layer remaining ratios \textit{without fine-tuning} to indicate their effectiveness in identifying important layers

\textit{Observations and analysis.} As shown in Tab.~\ref{tab:dp_ablation}, we observe that our DP-based strategy outperforms the other two strategies on both calibration datasets and metrics, especially under small remaining ratios, e.g., a  +9.4\% MMLU accuracy and a -33.1 PPL over the strongest baseline when remaining 18 layers.
This demonstrates the superiority of our method over the two baselines in selecting important layers that optimize the target calibration metric, thus significantly contributing to knowledge preservation.

\textit{Remark.} This set of experiments supports our analysis in Sec.~\ref{sec:method_selection} that the superiority of our method arises from its consideration of different layers' joint contributions, rather than focusing on single-layer importance~\cite{kim2024shortened}, and its avoidance of reliance on pre-defined rules~\cite{gromov2024unreasonable}, thus ensuring generality.

\begin{table}[!t]
\centering
\caption{Ablation Study on the effectiveness of the DP-based depth shrinking on LLaMA2 7B.}\label{tab:dp_ablation}
\vspace{-0.5em}
\resizebox{\linewidth}{!}{
\begin{tabular}{cccccccccccc}\toprule
\textbf{Calib. Data} & \textbf{Method} &\textbf{24} &\textbf{23} &\textbf{22} &\textbf{21} &\textbf{20} &\textbf{19} &\textbf{18} &\textbf{17} &\textbf{16} \\\cmidrule{1-11}
\multirow{3}{*}{Wikitext2 $\downarrow$} &Unreasonable~\cite{gromov2024unreasonable} &12.25 &13.72 &18.61 &31.35 &44.28 &57.34 &80.29 &120.57 &188.27 \\\cmidrule{2-11}
&ShortenLLaMA~\cite{kim2024shortened} &11.75 &13.18 &17.83 &29.61 &39.42 &53.27 &71.80 &106.63 &154.69 \\\cmidrule{2-11}
&\textbf{Ours} &\textbf{11.65} &\textbf{12.77} &\textbf{17.59} &\textbf{20.06} &\textbf{23.77} &\textbf{28.83} &\textbf{38.70} &\textbf{70.87} &\textbf{95.16} \\\cmidrule{1-11}
\multirow{3}{*}{MMLU (\%) $\uparrow$} &Unreasonable~\cite{gromov2024unreasonable} &40.8 &39.5 &37.5 &33.5 &34.5 &34.0 &30.3 &30.6 &25.8 \\\cmidrule{2-11}
&ShortenLLaMA~\cite{kim2024shortened} &42.0 &34.7 &35.1 &32.1 &32.0 &33.5 &33.7 &29.3 &26.6 \\\cmidrule{2-11}
&\textbf{Ours} &\textbf{46.2} &\textbf{44.8} &\textbf{44.6} &\textbf{44.1} &\textbf{41.2} &\textbf{41.3} &\textbf{43.1} &\textbf{34.7} &\textbf{28.9} \\
\bottomrule
\end{tabular}
}
\end{table}

\begin{table}[!t]
\centering
\caption{Ablation Study on different components in our \METHOD{} on LLaMA2 7B.}\label{tab:component}
\vspace{-0.5em}
\resizebox{\linewidth}{!}{
\begin{tabular}{ccccccc}\toprule
\multirow{2}{*}{\textbf{Method}} &\multicolumn{2}{c}{\textbf{32}} &\multicolumn{2}{c}{\textbf{24}} &\multicolumn{2}{c}{\textbf{20}}\\\cmidrule{2-7}
&\textbf{Wikitext2} $\downarrow$ &\textbf{MMLU (\%)} $\uparrow$ &\textbf{Wikitext2} $\downarrow$ &\textbf{MMLU (\%)} $\uparrow$ &\textbf{Wikitext2} $\downarrow$ &\textbf{MMLU (\%)} $\uparrow$ \\\cmidrule{1-7}
Per-subnet ft. &\textbf{5.54} &46.4 &\textbf{10.57} &41.9 &\textbf{15.94} &\textbf{41.7} \\\midrule 
- SMoL (+full model) & 5.82 & 46.6 & 38.48 & 32.6 & 167.74 & 36.4 \\
- SMoL (+LoRA) &6.97 &40.6 &12.71 &40.0 &19.12 &37.9 \\
- Loss-mag. Balancing &6.77 &42.0 &12.63 &40.1 &18.19 &39.3\\ \midrule
Full (\METHOD{}) &6.36 &\textbf{47.2} &12.40 &\textbf{45.1} &18.15 &41.0 \\
\bottomrule
\end{tabular}
}
\vspace{-0.5em}
\end{table}

\textbf{The SMoL adapter.} To assess the efficacy of our SMoL adapter, we substitute it with full model fine-tuning or the standard LoRA~\cite{hu2021lora} and benchmark it against our \METHOD{} with SMoL as well as the per-subnet fine-tuning variant. To more clearly demonstrate its efficacy, this experiment is conducted under a depth-shrinking-only setting, which only enables depth shrinking ranging from 20 to 32 layers during one-for-all fine-tuning. We select three layer configurations, covering both the largest and smallest ones, to report the performance.

\textit{Observations and analysis.}
As illustrated in Tab.~\ref{tab:component}, we observe that \underline{(1)} full model fine-tuning on a relatively small corpus results in suboptimal performance with large PPL; \underline{(2)} employing the standard LoRA stabilizes the one-for-all fine-tuning compared to full model fine-tuning, albeit with notable performance reductions in larger subnets, e.g., a 5.8\% MMLU accuracy drop in the largest subnet compared to per-subnet fine-tuning; \underline{(3)} when equipped with our SMoL, the MMLU accuracy of all subnets is substantially enhanced, e.g., a +6.6\% improvement on the largest subnet compared to the LoRA case, even exceeding that of the per-subnet counterparts. This demonstrates the essential capability of our \METHOD{} to deliver high-quality LLM subnets across a wide range of accuracy-efficiency trade-offs.

\textbf{The loss-magnitude balancing scheme.}
As shown in Tab.~\ref{tab:component}, we further disable the loss-magnitude balancing scheme and observe a significant MMLU accuracy drop in larger subnets. This supports our analysis in Sec.~\ref{sec:method_objective} that the larger losses from smaller subnets may dominate the overall objective, thereby impairing the fine-tuning of larger subnets and underscoring the necessity of our method.

\vspace{-0.3em}
\subsection{Ablation Study: The Selection of Calibration Datasets}
\label{sec:exp_calibration_dataset}
\vspace{-0.3em}

We conduct an ablation study on the choice of calibration datasets for our DP-based depth shrinking introduced in Sec.~\ref{sec:method_selection}. Specifically, we use accuracy on the training set of MMLU~\cite{hendrycks2020measuring} and PPL on the training sets of Wikitext2~\cite{wikitext2}/BookCorpus~\cite{zhu2015aligning} as target metrics. We report the evaluation metrics, including MMLU test accuracy and Wikitext2 test PPL, under different layer remaining ratios both after calibration and after one-for-all fine-tuning under a depth-shrinking-only setting.

\begin{table}[!t]
\centering
\caption{Ablation Study on the selection of calibration datasets on LLaMA2 7B.}\label{tab:calibration_dataset}
\resizebox{\linewidth}{!}{
\begin{tabular}{ccccccc}\toprule
\multirow{2}{*}{\textbf{Calib. Data}} &\multicolumn{2}{c}{\textbf{32}} &\multicolumn{2}{c}{\textbf{24}} &\multicolumn{2}{c}{\textbf{20}}\\\cmidrule{2-7}
 &\textbf{Wikitext2} $\downarrow$ &\textbf{MMLU (\%)} $\uparrow$ &\textbf{Wikitext2} $\downarrow$ &\textbf{MMLU (\%)} $\uparrow$ &\textbf{Wikitext2} $\downarrow$ &\textbf{MMLU (\%)} $\uparrow$ \\\midrule
BookCorpus &5.64 &46.4 &16.52 &26.8 &31.70 &27.0 \\
Wikitext2 &5.64 &46.4 &\textbf{11.92} &26.4 &\textbf{23.77} &24.4 \\
MMLU &5.64 &46.4 &293.01 &\textbf{45.8} &1338.68 &\textbf{41.2} \\\midrule \midrule
BookCorpus (ft) &6.55 &45.4 &10.55 & 29.7 &14.36 &25.7 \\
Wikitext2 (ft) &6.41 &\textbf{43.4} &\textbf{9.58} &32.6 &\textbf{13.26} &23.6 \\
MMLU (ft) &\textbf{6.36} &47.2 &12.40 & \textbf{45.1} &18.15 &\textbf{41.0} \\
\bottomrule
\end{tabular}
}
\vspace{-1.5em}
\end{table}

\textit{Observations.} As shown in Tab.~\ref{tab:calibration_dataset}, we observe that \underline{(1)} after calibration without fine-tuning, the subnets perform well on the evaluation metric for which they were calibrated and underperform in terms of the other metric; \underline{(2)} after fine-tuning, the subnets calibrated using PPL continue to perform poorly on MMLU accuracy, indicating a severe loss of factual knowledge. In contrast, the subnets calibrated using MMLU accuracy achieve notably lower PPL compared to before fine-tuning, even on par with the subnets calibrated using PPL, while still maintaining high MMLU accuracy.

\textit{The key insight.} This set of experiments indicates that the loss of factual knowledge during compression is hard to restore during fine-tuning, echoing the observations in~\cite{gekhman2024does}, while the language modeling capability is easier to recover through fine-tuning. As such, we adopted MMLU as the calibration dataset throughout the previous experiments, and we believe this insight could inspire future LLM compression frameworks and calibration metrics.

\vspace{-0.3em}
\subsection{Limitations and Future Work}
\label{sec:limitations}
\vspace{-0.3em}

One limitation of our work is that due to the limited fine-tuning data and resources, our methodology is applied to parameter-efficient fine-tuning, which mitigates gradient conflicts under small data conditions while limiting the achievable accuracy-efficiency trade-off. We anticipate that by leveraging more extensive fine-tuning data beyond our current use of Alpaca~\cite{alpaca} and extending our design insights regarding subnet selection and gradient conflict mitigation, more aggressive accuracy-efficiency trade-offs can be achieved, which will be the focus of our future work.
\section{Related Work}
\label{sec:related_work}

\textbf{Large language models.}
Before the advent of LLMs, transformer-based language models~\cite{vaswani2017attention,devlin2018bert,2020t5,roberts2022t5x} demonstrated their ability to effectively analyze relationships among tokens in complex input sequences, facilitated by the attention mechanism~\cite{vaswani2017attention}. These models also exhibit notable scalability~\cite{qin2023scaling,kaplan2020scaling,biderman2023pythia} with respect to model size and the scale of pre-training datasets. This decent scalability has led to the emergence of LLMs, such as GLM~\cite{du2021glm}, OPT~\cite{zhang2022opt}, BLOOM~\cite{workshop2022bloom}, the Llama family~\cite{touvron2023llama,touvron2023llama,llama3}, Gemma~\cite{team2024gemma}, and GPT-4~\cite{achiam2023gpt}, which exhibit impressive zero-shot and few-shot in-context learning capabilities. However, these LLMs often feature billions of parameters and prohibitive computation complexity, which limits their widespread use across diverse platforms.

\textbf{Large language model compression.} To facilitate efficient deployment of LLMs in real-world applications, existing works primarily focus on compressing LLMs by extending traditional compression techniques such as knowledge distillation~\cite{fu2023specializing,hsieh2023distilling}, quantization~\cite{frantar2022optq,dettmers2022llmint8,xiao2022smoothquant,frantar2023gptq,dettmers2023spqr,lin2023awq}, system acceleration~\cite{dao2023flashattention,kwon2023efficient}, and pruning~\cite{sun2023simple,frantar2023sparsegpt,ma2023llmv3}. Our work is most closely related to LLM pruning. Along this direction, early works~\cite{frantar2023sparsegpt,sun2023simple} employ unstructured and semi-structured pruning~\cite{zhou2021learning} by zeroing out connections among neurons. Despite their plausible performance, these methods require specialized support to achieve real-device speedup. To benefit commodity platforms, structured LLM pruning methods remove more coarse-grained components, such as all connections related to a single neuron~\cite{ma2023llm,an2024fluctuation,xia2023sheared} or even entire layers~\cite{kim2024shortened,gromov2024unreasonable}. For instance, LLM-Pruner~\cite{ma2023llm} and FLAP~\cite{an2024fluctuation} reduce LLM width by eliminating identified redundant neurons, while Sheared-LLaMA~\cite{xia2023sheared} learns a set of binary masks to reduce both the width and depth of LLMs. However, these methods either focus on a single dimension of compression (i.e., depth or width) with limited efficiency improvements, or they require a costly fine-tuning process for each target configuration and platform. In contrast, our \METHOD{} can instantly extract subnets of arbitrary shapes that reach the accuracy-efficiency frontier, thus facilitating rapid deployment across devices.

\textbf{One-for-all networks.}
Slimmable networks~\cite{yu2018slimmable,yu2019universally} are pioneering works that enable a single model to operate at varying widths. Follow-up works~\cite{yu2020bignas,yu2019autoslim,wang2021attentivenas,wang2021alphanet,gong2022nasvit,cai2019once} further extend this approach to train more general one-for-all networks with switchable depth and width, thus enabling tasks like neural architecture search. In particular, BigNAS~\cite{yu2020bignas} builds one-for-all networks using a sandwich sampling strategy that samples a random set of subnets and jointly trains them in each iteration through in-place distillation, where the largest model guides the learning of the smaller ones. This approach has been inherited by subsequent works~\cite{wang2021attentivenas,wang2021alphanet,gong2022nasvit}. Additionally, this idea has been extended to any-precision networks~\cite{jin2020adabits,guerra2020switchable,yu2021any} that allow switchable precision at runtime. 
A very recent work~\cite{park2024any} has further extended this concept to any-precision LLMs.

Nevertheless, directly applying these methods to the depth and width of pre-trained LLMs would likely fail because their subnet sampling strategies often select the first layers of a model or the first channels of a layer, which are intended for models trained from scratch. This approach is unsuitable for pre-trained LLMs, as it may omit layers or neurons containing crucial knowledge. Our \METHOD{} framework addresses these challenges by developing three key components: the subnet selection strategy, the trainable adapter design, and the fine-tuning objective.
One concurrent work~\cite{cai2024flextron} also aims to train many-in-one LLMs that support instant subnet derivation, targeting a full-model continual training setting on 90 billion tokens. In contrast, our method targets a parameter-efficient tuning setting with only 8.5 million fine-tuning tokens.
\section{Conclusion}
\label{sec:conclusion}

In this work, we present a framework called \METHOD{} that grants a given LLM the capability to instantly deliver subnets of arbitrary shapes, which achieve the accuracy-efficiency frontier and can be extracted immediately after a one-time fine-tuning. This is achieved by the development of three dedicated components that enable \METHOD{}'s one-for-all fine-tuning scheme: a knowledge-preserving subnet selection strategy, an SMoL adapter, and an in-place distillation objective with loss-magnitude balancing. Extensive experiments validate that our \METHOD{} framework can deliver efficient LLMs with instantly serviceable subnets of any shape, which outperform SOTA LLM compression techniques in terms of the accuracy-efficiency trade-off.
We believe this work is promising to facilitate the wider use of existing and emerging public LLMs by making them instantly deployable on varying platforms and applications, and by providing a new perspective on efficient LLM deployment, thus inspiring future solutions.

\section*{Acknowledgement}
The work is supported by an National Science Foundation (NSF) CAREER award (Award number: 2345577) and CoCoSys, one of the seven centers in JUMP 2.0, a Semiconductor Research Corporation (SRC) program sponsored by DARPA.

\bibliographystyle{unsrt}
\bibliography{ref}

\newpage
\section*{NeurIPS Paper Checklist}

\begin{enumerate}

\item {\bf Claims}
    \item[] Question: Do the main claims made in the abstract and introduction accurately reflect the paper's contributions and scope?
    \item[] Answer: \answerYes{}
    \item[] Justification: We have accurately summarized our paper's contributions and scope in the abstract and introduction.
    \item[] Guidelines:
    \begin{itemize}
        \item The answer NA means that the abstract and introduction do not include the claims made in the paper.
        \item The abstract and/or introduction should clearly state the claims made, including the contributions made in the paper and important assumptions and limitations. A No or NA answer to this question will not be perceived well by the reviewers. 
        \item The claims made should match theoretical and experimental results, and reflect how much the results can be expected to generalize to other settings. 
        \item It is fine to include aspirational goals as motivation as long as it is clear that these goals are not attained by the paper. 
    \end{itemize}

\item {\bf Limitations}
    \item[] Question: Does the paper discuss the limitations of the work performed by the authors?
    \item[] Answer: \answerYes{}
    \item[] Justification: We discussed this in Sec. 5.5 of our paper.
    \item[] Guidelines:
    \begin{itemize}
        \item The answer NA means that the paper has no limitation while the answer No means that the paper has limitations, but those are not discussed in the paper. 
        \item The authors are encouraged to create a separate "Limitations" section in their paper.
        \item The paper should point out any strong assumptions and how robust the results are to violations of these assumptions (e.g., independence assumptions, noiseless settings, model well-specification, asymptotic approximations only holding locally). The authors should reflect on how these assumptions might be violated in practice and what the implications would be.
        \item The authors should reflect on the scope of the claims made, e.g., if the approach was only tested on a few datasets or with a few runs. In general, empirical results often depend on implicit assumptions, which should be articulated.
        \item The authors should reflect on the factors that influence the performance of the approach. For example, a facial recognition algorithm may perform poorly when image resolution is low or images are taken in low lighting. Or a speech-to-text system might not be used reliably to provide closed captions for online lectures because it fails to handle technical jargon.
        \item The authors should discuss the computational efficiency of the proposed algorithms and how they scale with dataset size.
        \item If applicable, the authors should discuss possible limitations of their approach to address problems of privacy and fairness.
        \item While the authors might fear that complete honesty about limitations might be used by reviewers as grounds for rejection, a worse outcome might be that reviewers discover limitations that aren't acknowledged in the paper. The authors should use their best judgment and recognize that individual actions in favor of transparency play an important role in developing norms that preserve the integrity of the community. Reviewers will be specifically instructed to not penalize honesty concerning limitations.
    \end{itemize}

\item {\bf Theory Assumptions and Proofs}
    \item[] Question: For each theoretical result, does the paper provide the full set of assumptions and a complete (and correct) proof?
    \item[] Answer: \answerNA{} 
    \item[] Justification: We didn't provide theoretical results.
    \item[] Guidelines:
    \begin{itemize}
        \item The answer NA means that the paper does not include theoretical results. 
        \item All the theorems, formulas, and proofs in the paper should be numbered and cross-referenced.
        \item All assumptions should be clearly stated or referenced in the statement of any theorems.
        \item The proofs can either appear in the main paper or the supplemental material, but if they appear in the supplemental material, the authors are encouraged to provide a short proof sketch to provide intuition. 
        \item Inversely, any informal proof provided in the core of the paper should be complemented by formal proofs provided in appendix or supplemental material.
        \item Theorems and Lemmas that the proof relies upon should be properly referenced. 
    \end{itemize}

    \item {\bf Experimental Result Reproducibility}
    \item[] Question: Does the paper fully disclose all the information needed to reproduce the main experimental results of the paper to the extent that it affects the main claims and/or conclusions of the paper (regardless of whether the code and data are provided or not)?
    \item[] Answer: \answerYes{} 
    \item[] Justification: Yes, we have elaborated on our settings and provided the code.
    \item[] Guidelines:
    \begin{itemize}
        \item The answer NA means that the paper does not include experiments.
        \item If the paper includes experiments, a No answer to this question will not be perceived well by the reviewers: Making the paper reproducible is important, regardless of whether the code and data are provided or not.
        \item If the contribution is a dataset and/or model, the authors should describe the steps taken to make their results reproducible or verifiable. 
        \item Depending on the contribution, reproducibility can be accomplished in various ways. For example, if the contribution is a novel architecture, describing the architecture fully might suffice, or if the contribution is a specific model and empirical evaluation, it may be necessary to either make it possible for others to replicate the model with the same dataset, or provide access to the model. In general. releasing code and data is often one good way to accomplish this, but reproducibility can also be provided via detailed instructions for how to replicate the results, access to a hosted model (e.g., in the case of a large language model), releasing of a model checkpoint, or other means that are appropriate to the research performed.
        \item While NeurIPS does not require releasing code, the conference does require all submissions to provide some reasonable avenue for reproducibility, which may depend on the nature of the contribution. For example
        \begin{enumerate}
            \item If the contribution is primarily a new algorithm, the paper should make it clear how to reproduce that algorithm.
            \item If the contribution is primarily a new model architecture, the paper should describe the architecture clearly and fully.
            \item If the contribution is a new model (e.g., a large language model), then there should either be a way to access this model for reproducing the results or a way to reproduce the model (e.g., with an open-source dataset or instructions for how to construct the dataset).
            \item We recognize that reproducibility may be tricky in some cases, in which case authors are welcome to describe the particular way they provide for reproducibility. In the case of closed-source models, it may be that access to the model is limited in some way (e.g., to registered users), but it should be possible for other researchers to have some path to reproducing or verifying the results.
        \end{enumerate}
    \end{itemize}

\item {\bf Open access to data and code}
    \item[] Question: Does the paper provide open access to the data and code, with sufficient instructions to faithfully reproduce the main experimental results, as described in supplemental material?
    \item[] Answer: \answerYes{} 
    \item[] Justification: We have open-sourced our code.
    \item[] Guidelines:
    \begin{itemize}
        \item The answer NA means that paper does not include experiments requiring code.
        \item Please see the NeurIPS code and data submission guidelines (\url{https://nips.cc/public/guides/CodeSubmissionPolicy}) for more details.
        \item While we encourage the release of code and data, we understand that this might not be possible, so “No” is an acceptable answer. Papers cannot be rejected simply for not including code, unless this is central to the contribution (e.g., for a new open-source benchmark).
        \item The instructions should contain the exact command and environment needed to run to reproduce the results. See the NeurIPS code and data submission guidelines (\url{https://nips.cc/public/guides/CodeSubmissionPolicy}) for more details.
        \item The authors should provide instructions on data access and preparation, including how to access the raw data, preprocessed data, intermediate data, and generated data, etc.
        \item The authors should provide scripts to reproduce all experimental results for the new proposed method and baselines. If only a subset of experiments are reproducible, they should state which ones are omitted from the script and why.
        \item At submission time, to preserve anonymity, the authors should release anonymized versions (if applicable).
        \item Providing as much information as possible in supplemental material (appended to the paper) is recommended, but including URLs to data and code is permitted.
    \end{itemize}

\item {\bf Experimental Setting/Details}
    \item[] Question: Does the paper specify all the training and test details (e.g., data splits, hyperparameters, how they were chosen, type of optimizer, etc.) necessary to understand the results?
    \item[] Answer: \answerYes{} 
    \item[] Justification: We offered this information in Sec. 5.1 of our paper and also provided sufficient references.
    \item[] Guidelines:
    \begin{itemize}
        \item The answer NA means that the paper does not include experiments.
        \item The experimental setting should be presented in the core of the paper to a level of detail that is necessary to appreciate the results and make sense of them.
        \item The full details can be provided either with the code, in appendix, or as supplemental material.
    \end{itemize}

\item {\bf Experiment Statistical Significance}
    \item[] Question: Does the paper report error bars suitably and correctly defined or other appropriate information about the statistical significance of the experiments?
    \item[] Answer: \answerNo{} 
    \item[] Justification: We follow our baselines and do not report the error bars as the results are relatively stable.
    \item[] Guidelines:
    \begin{itemize}
        \item The answer NA means that the paper does not include experiments.
        \item The authors should answer "Yes" if the results are accompanied by error bars, confidence intervals, or statistical significance tests, at least for the experiments that support the main claims of the paper.
        \item The factors of variability that the error bars are capturing should be clearly stated (for example, train/test split, initialization, random drawing of some parameter, or overall run with given experimental conditions).
        \item The method for calculating the error bars should be explained (closed form formula, call to a library function, bootstrap, etc.)
        \item The assumptions made should be given (e.g., Normally distributed errors).
        \item It should be clear whether the error bar is the standard deviation or the standard error of the mean.
        \item It is OK to report 1-sigma error bars, but one should state it. The authors should preferably report a 2-sigma error bar than state that they have a 96\% CI, if the hypothesis of Normality of errors is not verified.
        \item For asymmetric distributions, the authors should be careful not to show in tables or figures symmetric error bars that would yield results that are out of range (e.g. negative error rates).
        \item If error bars are reported in tables or plots, The authors should explain in the text how they were calculated and reference the corresponding figures or tables in the text.
    \end{itemize}

\item {\bf Experiments Compute Resources}
    \item[] Question: For each experiment, does the paper provide sufficient information on the computer resources (type of compute workers, memory, time of execution) needed to reproduce the experiments?
    \item[] Answer: \answerYes{} 
    \item[] Justification: We provided this information in Sec. 5.1 of our paper.
    \item[] Guidelines:
    \begin{itemize}
        \item The answer NA means that the paper does not include experiments.
        \item The paper should indicate the type of compute workers CPU or GPU, internal cluster, or cloud provider, including relevant memory and storage.
        \item The paper should provide the amount of compute required for each of the individual experimental runs as well as estimate the total compute. 
        \item The paper should disclose whether the full research project required more compute than the experiments reported in the paper (e.g., preliminary or failed experiments that didn't make it into the paper). 
    \end{itemize}
    
\item {\bf Code Of Ethics}
    \item[] Question: Does the research conducted in the paper conform, in every respect, with the NeurIPS Code of Ethics \url{https://neurips.cc/public/EthicsGuidelines}?
    \item[] Answer: \answerYes{} 
    \item[] Justification: Yes, it conforms with NeurIPS Code of Ethnics.
    \item[] Guidelines:
    \begin{itemize}
        \item The answer NA means that the authors have not reviewed the NeurIPS Code of Ethics.
        \item If the authors answer No, they should explain the special circumstances that require a deviation from the Code of Ethics.
        \item The authors should make sure to preserve anonymity (e.g., if there is a special consideration due to laws or regulations in their jurisdiction).
    \end{itemize}

\item {\bf Broader Impacts}
    \item[] Question: Does the paper discuss both potential positive societal impacts and negative societal impacts of the work performed?
    \item[] Answer: \answerNA{} 
    \item[] Justification: This work targets the acceleration of large language models to facilitate its widespread use and thus does not suffer from obvious negative societal impact.
    \item[] Guidelines:
    \begin{itemize}
        \item The answer NA means that there is no societal impact of the work performed.
        \item If the authors answer NA or No, they should explain why their work has no societal impact or why the paper does not address societal impact.
        \item Examples of negative societal impacts include potential malicious or unintended uses (e.g., disinformation, generating fake profiles, surveillance), fairness considerations (e.g., deployment of technologies that could make decisions that unfairly impact specific groups), privacy considerations, and security considerations.
        \item The conference expects that many papers will be foundational research and not tied to particular applications, let alone deployments. However, if there is a direct path to any negative applications, the authors should point it out. For example, it is legitimate to point out that an improvement in the quality of generative models could be used to generate deepfakes for disinformation. On the other hand, it is not needed to point out that a generic algorithm for optimizing neural networks could enable people to train models that generate Deepfakes faster.
        \item The authors should consider possible harms that could arise when the technology is being used as intended and functioning correctly, harms that could arise when the technology is being used as intended but gives incorrect results, and harms following from (intentional or unintentional) misuse of the technology.
        \item If there are negative societal impacts, the authors could also discuss possible mitigation strategies (e.g., gated release of models, providing defenses in addition to attacks, mechanisms for monitoring misuse, mechanisms to monitor how a system learns from feedback over time, improving the efficiency and accessibility of ML).
    \end{itemize}
    
\item {\bf Safeguards}
    \item[] Question: Does the paper describe safeguards that have been put in place for responsible release of data or models that have a high risk for misuse (e.g., pretrained language models, image generators, or scraped datasets)?
    \item[] Answer: \answerNA{} 
    \item[] Justification: We built on top of public models and datasets and do not suffer from high risks of misuse.
    \item[] Guidelines:
    \begin{itemize}
        \item The answer NA means that the paper poses no such risks.
        \item Released models that have a high risk for misuse or dual-use should be released with necessary safeguards to allow for controlled use of the model, for example by requiring that users adhere to usage guidelines or restrictions to access the model or implementing safety filters. 
        \item Datasets that have been scraped from the Internet could pose safety risks. The authors should describe how they avoided releasing unsafe images.
        \item We recognize that providing effective safeguards is challenging, and many papers do not require this, but we encourage authors to take this into account and make a best faith effort.
    \end{itemize}

\item {\bf Licenses for existing assets}
    \item[] Question: Are the creators or original owners of assets (e.g., code, data, models), used in the paper, properly credited and are the license and terms of use explicitly mentioned and properly respected?
    \item[] Answer: \answerYes{} 
    \item[] Justification: Yes, we have added the reference for all used data.
    \item[] Guidelines:
    \begin{itemize}
        \item The answer NA means that the paper does not use existing assets.
        \item The authors should cite the original paper that produced the code package or dataset.
        \item The authors should state which version of the asset is used and, if possible, include a URL.
        \item The name of the license (e.g., CC-BY 4.0) should be included for each asset.
        \item For scraped data from a particular source (e.g., website), the copyright and terms of service of that source should be provided.
        \item If assets are released, the license, copyright information, and terms of use in the package should be provided. For popular datasets, \url{paperswithcode.com/datasets} has curated licenses for some datasets. Their licensing guide can help determine the license of a dataset.
        \item For existing datasets that are re-packaged, both the original license and the license of the derived asset (if it has changed) should be provided.
        \item If this information is not available online, the authors are encouraged to reach out to the asset's creators.
    \end{itemize}

\item {\bf New Assets}
    \item[] Question: Are new assets introduced in the paper well documented and is the documentation provided alongside the assets?
    \item[] Answer: \answerNA{} 
    \item[] Justification: We did not introduce new assets.
    \item[] Guidelines:
    \begin{itemize}
        \item The answer NA means that the paper does not release new assets.
        \item Researchers should communicate the details of the dataset/code/model as part of their submissions via structured templates. This includes details about training, license, limitations, etc. 
        \item The paper should discuss whether and how consent was obtained from people whose asset is used.
        \item At submission time, remember to anonymize your assets (if applicable). You can either create an anonymized URL or include an anonymized zip file.
    \end{itemize}

\item {\bf Crowdsourcing and Research with Human Subjects}
    \item[] Question: For crowdsourcing experiments and research with human subjects, does the paper include the full text of instructions given to participants and screenshots, if applicable, as well as details about compensation (if any)? 
    \item[] Answer: \answerNA{} 
    \item[] Justification: We did not perform crowdsourcing experiments.
    \item[] Guidelines:
    \begin{itemize}
        \item The answer NA means that the paper does not involve crowdsourcing nor research with human subjects.
        \item Including this information in the supplemental material is fine, but if the main contribution of the paper involves human subjects, then as much detail as possible should be included in the main paper. 
        \item According to the NeurIPS Code of Ethics, workers involved in data collection, curation, or other labor should be paid at least the minimum wage in the country of the data collector. 
    \end{itemize}

\item {\bf Institutional Review Board (IRB) Approvals or Equivalent for Research with Human Subjects}
    \item[] Question: Does the paper describe potential risks incurred by study participants, whether such risks were disclosed to the subjects, and whether Institutional Review Board (IRB) approvals (or an equivalent approval/review based on the requirements of your country or institution) were obtained?
    \item[] Answer: \answerNA{} 
    \item[] Justification: Our work didn't involve human participants.
    \item[] Guidelines:
    \begin{itemize}
        \item The answer NA means that the paper does not involve crowdsourcing nor research with human subjects.
        \item Depending on the country in which research is conducted, IRB approval (or equivalent) may be required for any human subjects research. If you obtained IRB approval, you should clearly state this in the paper. 
        \item We recognize that the procedures for this may vary significantly between institutions and locations, and we expect authors to adhere to the NeurIPS Code of Ethics and the guidelines for their institution. 
        \item For initial submissions, do not include any information that would break anonymity (if applicable), such as the institution conducting the review.
    \end{itemize}

\end{enumerate}

\end{document}